\pdfoutput=1

\documentclass[11pt]{article}

\usepackage[final]{acl}

\usepackage{times}
\usepackage{latexsym}
\usepackage{booktabs}
\usepackage{algorithm}
\usepackage{tikz}
\usepackage{pgfplots}
\usepackage{algorithmic}
\usepackage{float}
\usepackage{caption}
\usepackage{subcaption}
\pgfplotsset{compat=1.17}
\usetikzlibrary{shapes.geometric}
\usepackage{subcaption}
\usepackage{soul}
\usepackage{url}
\usepackage{graphicx}
\usepackage{amsmath}
\usepackage{amsthm}
\usepackage{amsfonts}
\usepackage{amssymb}
\usepackage{multicol}
\usepackage{hhline}

\usepackage[T1]{fontenc}
\usepackage[utf8]{inputenc}
\usepackage{amsmath}
\usepackage{microtype}

\usepackage{inconsolata}

\usepackage{graphicx}
\usepackage{multirow}
\title{RASD: Retrieval-Augmented Speculative Decoding}

\author{
  \textbf{Guofeng Quan} 
  \textbf{Wenfeng Feng} 
  \textbf{Chuzhan Hao} 
  \textbf{Guochao Jiang} \\
  \textbf{Yuewei Zhang} 
  \textbf{Hao Wang} \\
  \textsuperscript{1}Alibaba Cloud, Alibaba Group \\
  \texttt{\{quanguofeng.qgf,wenfeng.fwf,hao.hcz,anyue.jgc,liyou.zyw\}}@alibaba-inc.com
  \\
  \texttt{cashenry@126.com}
}

\begin{document}
\maketitle

\begin{abstract}
Speculative decoding accelerates inference in large language models (LLMs) by generating draft tokens for target model verification. Current approaches for obtaining draft tokens rely on lightweight draft models or additional model structures to generate draft tokens and retrieve context from databases. Due to the draft model's small size and limited training data, model-based speculative decoding frequently becomes less effective in out-of-domain scenarios. Additionally, the time cost of the drafting phase results in a low upper limit on acceptance length during the verification step, limiting overall efficiency.
This paper proposes RASD (Retrieval-Augmented Speculative Decoding), which adopts retrieval methods to enhance model-based speculative decoding. We introduce tree pruning and tree fusion to achieve this. Specifically, we develop a pruning method based on the draft model's probability distribution to construct the optimal retrieval tree. Second, we employ the longest prefix matching algorithm to merge the tree generated by the draft model with the retrieval tree, resulting in a unified tree for verification.
Experimental results demonstrate that RASD achieves state-of-the-art inference acceleration across tasks such as DocQA, Summary, Code, and In-Domain QA. Moreover, RASD exhibits strong scalability, seamlessly integrating with various speculative decoding approaches, including both generation-based and retrieval-based methods.
\end{abstract}

\section{Introduction}

Transformer-based Large Language Models (LLMs) \cite{DBLP:conf/nips/VaswaniSPUJGKP17,DBLP:conf/nips/BrownMRSKDNSSAA20} exhibit remarkable capabilities and are extensively applied across diverse domains. However, autoregressive generation in LLMs produces tokens sequentially, resulting in slow inference speeds. To address this issue, an innovative approach called Speculative Decoding has been introduced \cite{DBLP:journals/corr/abs-2302-01318,DBLP:journals/corr/abs-2305-09781}. Speculative decoding modifies the inference process by dividing the LLM's task into two phases: a cost-efficient draft phase and a parallel verification phase. This strategy significantly improves computational parallelism in LLM inference. By enabling LLMs to generate multiple tokens simultaneously and minimizing the time spent in the drafting and verification phases, speculative decoding reduces the overall inference time \cite{DBLP:conf/icml/LeviathanKM23}.

\begin{figure}
    \centering
    \includegraphics[width=1\linewidth]{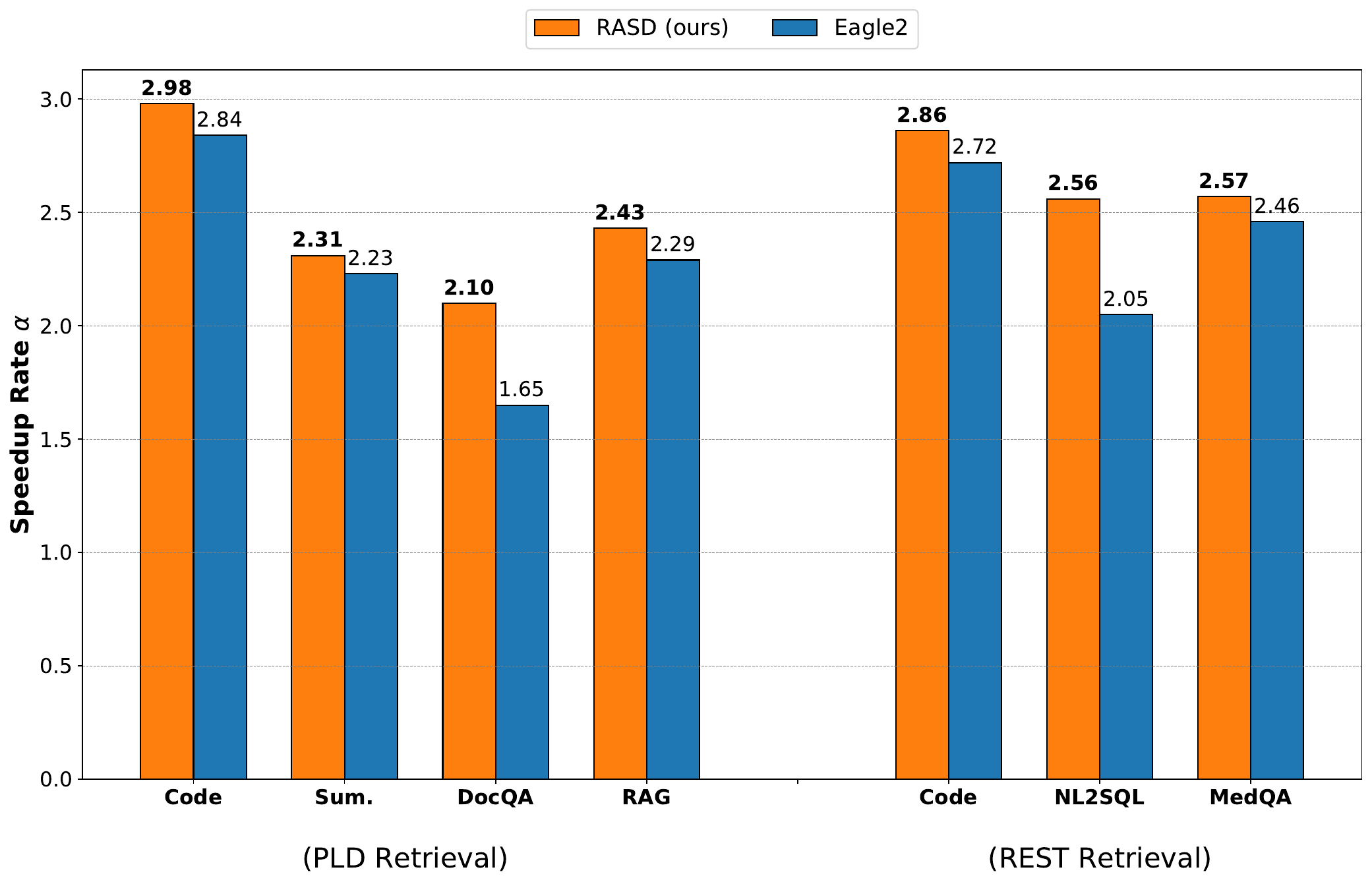}
    \caption{Speedup Performance of EAGLE2 vs. RASD with PLD and REST Retrieval Methods on Qwen2.5-14B.}
    \label{fig:speedup_rate}
\end{figure}

In speculative decoding, if the majority of draft tokens are correct, the overall decoding steps can be significantly reduced. Therefore, obtaining draft tokens with a high acceptance rate is essential.
Some researchers employ small draft models to predict draft tokens \cite{DBLP:conf/icml/LeviathanKM23,DBLP:journals/corr/abs-2302-01318,DBLP:conf/icml/LiW0024,DBLP:journals/corr/abs-2406-16858}. These draft models are trained to match the distribution of the target model and then generate draft tokens.
Other researchers have proposed using a parameter-efficient model structure to generate the next $k$ candidate tokens in a single forward pass of the target model \cite{DBLP:conf/icml/CaiLGPLCD24,DBLP:journals/corr/abs-2405-00263}. This approach also requires training the additional structure.
An alternative approach is retrieval-based speculative decoding, which is a train-free method \cite{saxena2023prompt,DBLP:conf/naacl/0012ZCL024}. Here, a retrieval library pre-defines tokens to follow the suffix of the current content as draft tokens. This approach is fast and eliminates the need for additional model training. Although it is generally less effective than generation-based methods, it performs well in specific knowledge-intensive scenarios.

In draft model-based speculative decoding, the acceptance rate of draft tokens hinges on the draft model's capabilities. However, these models typically have simple structures with limited parameters, which restricts their ability to effectively retain training data. Typically, a draft model is trained on a general dataset to perform well in common scenarios, such as the ShareGPT dataset. As a result, in specific scenarios, the draft model's token acceptance rate tends to be low. This issue arises because the draft model \textbf{lacks capabilities for out-of-domain scenarios}. There has been work on improving the generalizability of the draft model through feature-level supervision, which closes the representation gap between the draft model and the target model without requiring large amounts of data \cite{DBLP:conf/icml/LiW0024, du2024glide}. However, we have evaluated these methods on our domain datasets and discover that the draft model still does not perform well in the specific downstream domain after our test of business data.
Moreover, generating the draft token sequence requires multiple forward passes through the draft model, making the process time-intensive. Consequently, previous methods limit the length of candidate draft tokens. This restriction leads to a \textbf{low upper limit on the acceptance length} during a single verification step.

In this paper, we propose the RASD (Retrieval-Augmented Speculative Decoding) method to address the draft model's shortcomings mentioned above. First, we design an algorithm to select suitable retrieval results and then combine the draft tokens using a tree-fusion approach. This enables the generation of draft tokens that incorporate information from the language model and retrieval. Finally, the target model verifies the draft tokens to achieve acceleration.

We tested two retrieval methods PLD \cite{saxena2023prompt} and REST \cite{DBLP:conf/naacl/0012ZCL024} to enhance model-based speculative decoding. As shown in Figure \ref{fig:speedup_rate}, the PLD method performs well in tasks where the output includes input content. Therefore, we conducted comprehensive experiments using RASD (PLD) on datasets with these characteristics across four tasks: code generation, document question answering, summarization, and retrieval-augmented generation. We have conducted experiments on HumanEval \cite{DBLP:journals/corr/abs-2107-03374}, CNN/Daily Mail \cite{DBLP:conf/conll/NallapatiZSGX16}, MFQA \cite{DBLP:conf/acl/BaiLZL0HDLZHDTL24}, and DPR \cite{DBLP:conf/emnlp/KarpukhinOMLWEC20}. RASD (PLD) achieved the best results in all four tasks.

To extend the applicability of the RASD method, we also tested the REST retrieval method. REST is theoretically beneficial for any dataset when the database's data distribution closely matches the test set, excelling in knowledge-intensive tasks. We experimented with RASD (REST) on three tasks: HumanEval, MedQA \footnote{https://huggingface.co/datasets/lavita/medical-qa-datasets}, and NL2SQL. The results demonstrate that RASD (REST) effectively improves the acceleration ratio of speculative decoding.

\section{Related Work}
\subsection{EAGLE}
Model-based speculative decoding is widely regarded as the most effective approach for achieving acceleration. As the current state-of-the-art method, the EAGLE series method \cite{DBLP:conf/icml/LiW0024,DBLP:journals/corr/abs-2406-16858,DBLP:journals/corr/abs-2412-12639} is designed to provide feature supervision signals. The core structure of the EAGLE draft model includes one layer of the target model, sharing parameters with the embedding layer and the language model head. During training, only the parameters of this layer are updated, while the embedding layer and the language model head remain frozen. 

Before training, EAGLE processes the training data through the target model to extract the output features from its last layer. During draft model training, EAGLE combines the current embedding input with the previous output features of the target model to form a new input. It introduces a loss function based on the output features of the last layer, aiming to closely align the draft model’s output features with those of the target model.

EAGLE-2 \cite{DBLP:journals/corr/abs-2406-16858} follows the same design but replaces the static tree structure with dynamic drafting structures during decoding to generate higher-quality candidate trees. 

In RASD, we employ EAGLE-2 as the model-based speculative decoding method. Any speculative decoding approach that utilizes tree attention in drafting and verifying can be enhanced with RASD, as we do not change the method itself.

Before training, EAGLE processes the training data through the target model to extract the output features from its last layer. During draft model training, EAGLE combines the current embedding input with the previous output features of the target model to form a new input. It introduces a loss function based on the output features of the last layer, aiming to closely align the draft model's output features with those of the target model.
EAGLE-2 \cite{DBLP:journals/corr/abs-2406-16858} follows the same design but replaces the static tree structure with dynamic drafting structures during decoding to generate higher-quality candidate trees.
In RASD, we employ EAGLE-2 as the model-based speculative decoding method. Any speculative decoding approach that utilizes tree attention in drafting and verifying can be enhanced with RASD, as we do not change the method itself.

\subsection{PLD}
Retrieval-based speculative sampling methods perform well in knowledge-intensive scenarios. PLD (Prompt Lookup Decoding) is a simple and low-cost method for retrieving sequences from the input. However, it cannot predict new tokens or their combinations. It relies on the last $n$ tokens of the input for $n$-gram matching.
The original implementation\footnote{\url{https://github.com/apoorvumang/prompt-lookup-decoding/}} terminates at the first match. In our version, we return the first $k$ matching results instead.

\subsection{REST}
REST is a novel algorithm that leverages retrieval to generate draft tokens. Unlike PLD, which retrieves directly from the input, REST retrieves from a pre-defined context database. It utilizes existing knowledge, fetching relevant tokens based on the current context without relying on draft models. REST converts existing corpora into a retrieval library and organizes the results into a draft tree for verification by the target model. As a plug-and-play speculative sampling method, REST does not require additional draft model structures. While user-friendly, it is less effective compared to draft model-based methods.

We employ REST as the retrieval method to demonstrate the excellent scalability of RASD. All speculative decoding methods with external retrieval can be integrated into RASD.

\section{Retrieval-Augmented Speculative Decoding}
In this section, we first present the background of speculative decoding and then introduce our proposed RASD framework.
\subsection{Background: Speculative Decoding}
Speculative decoding with the draft model combines drafting and verification processes. In a single step of drafting, we use \(s\) to denote the input, which contains \(n\) tokens, including those from user input and tokens generated by previous speculative decoding processes: 
\begin{align}
     s=(x_1, \ldots, x_{n-1}, x_n).
\end{align}
The target model generates the first token \( y_0 \) based on the target model probability distribution computed by \(s\):
\begin{align}
    y_0 &\sim P(x \mid s; \theta_{\text{target}}),
\end{align}
where \(\theta_{\text{target}}\) represents the parameters of the target model. If the current step is not the first step of speculative decoding, the token \( y_0 \) will will be generated together with other tokens in the verification phase.

The draft model concatenates \(y_0\) to \(s\) to get the input \(s^{\prime}\). 
\begin{align}
    s^{\prime}=(x_1, \ldots, x_{n-1}, x_n , y_0). 
\end{align}
Given the input, the draft model generates multiple candidate tokens by autoregressive decoding. Assuming the \(m\)-th token is the last token of the current draft phase:
\begin{equation}
    \hat{p}_m = P(x \mid s^{\prime}, \hat{y}_1, \ldots, \hat{y}_{m-1}; \theta_{\text{draft}}),
\end{equation}
\begin{equation}
    \hat{y}_m \sim \hat{p}_m,
\end{equation}
\begin{equation}
    \hat{s}^{\prime} = (s^{\prime}, \hat{y}_1, \ldots, \hat{y}_m),
\end{equation}
where \(\hat{p}_m\) represents the probability distribution of the \(m\)-th draft token, \(\hat{y}_m\) represents the $m$-th draft token. $\hat{s^{\prime}}$ represents the output of the draft model in this step, and \(\theta_{\text{draft}}\) represents the parameters of the draft model. 

In the verification process, the target model verifies $\hat{s^{\prime}}$ in a single forward pass. The target model probability of the $k$-th token is \(p_k\). The draft token $\hat{y}_k$ has an acceptance probability \(\min\left(1, \frac{p_k}{\hat{p}_k}\right)\). If the draft token \(\hat{y}_k\) is rejected, all subsequent tokens are discarded, and this token is resampled from a distribution \(\lVert \max(0, p_k- \hat{p}_k) \rVert\). If \(\hat{y}_k\) is accepted, \(\hat{y}_k\) is transferred to $y_k$ and the target model continues to verify the next token \(\hat{y}_{k+1}\) until the last token. 
Finally, the accepted sequence and the resampled token compose $s^{\prime}$ for the next speculative decoding step.
The probability distribution generated by the verification method is equivalent to sampling directly from the target model \cite{DBLP:conf/icml/LeviathanKM23}.

We employ tree attention for the simultaneous verification of multiple candidates. Traditional causal attention masks are designed for linear sequences, where each token attends to all previous tokens, limiting speculative decoding to verify only one sequence at a time. Tree attention alters the attention mask to compress multiple sequences into a single merged sequence while preserving a tree structure. Each child node attends only to its parent nodes. In summary, speculative decoding, through guess-and-verify and tree attention, robustly and efficiently improves inference latency compared to autoregressive decoding. RASD relies on tree attention to achieve retrieval-augmented speculative decoding.

\begin{figure*}[h!]
   \centering   \includegraphics[width=1\linewidth]{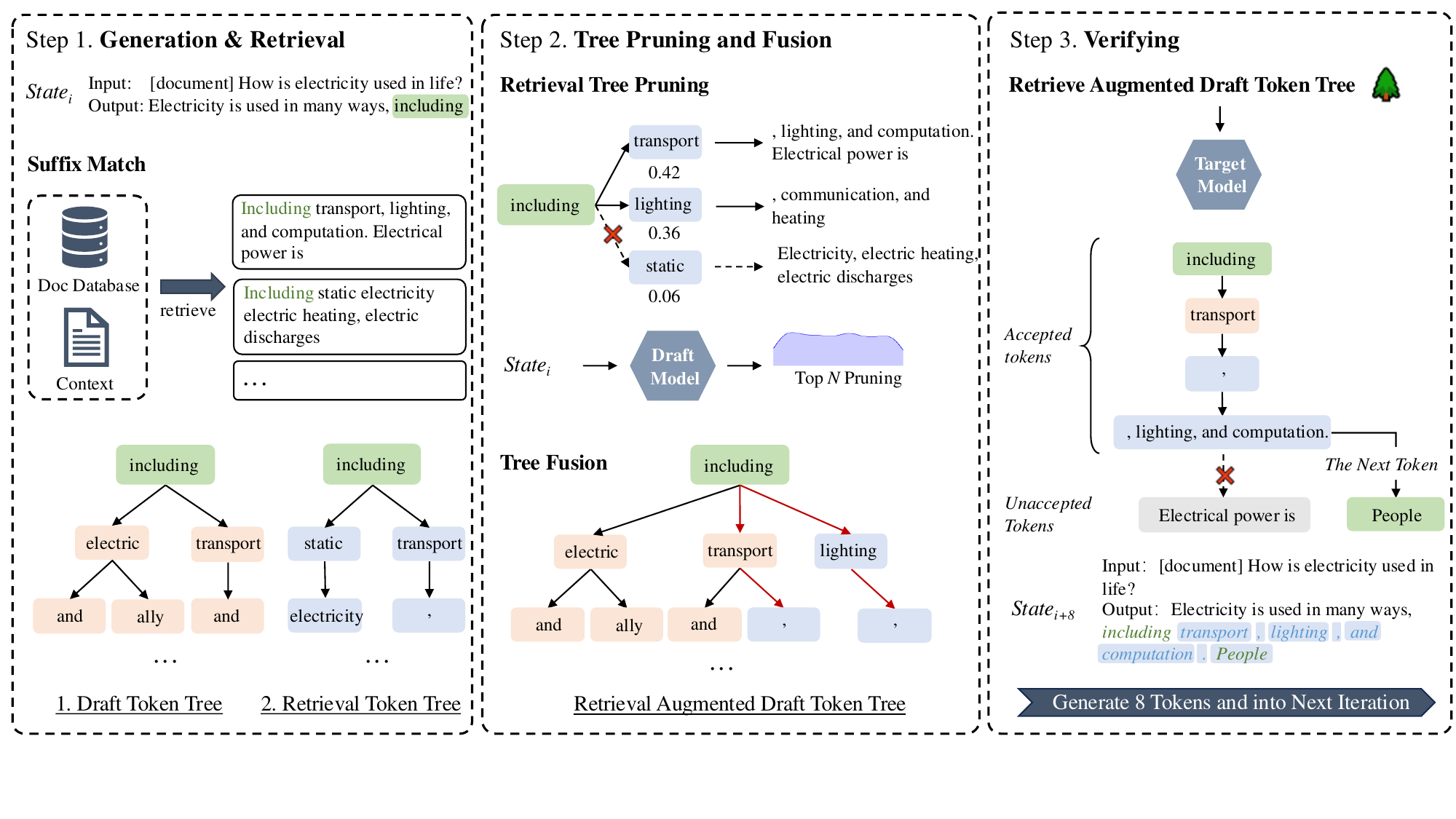}
   \caption{An overview of RASD. We obtain the draft token tree and retrieval results through the draft model generation and retrieval methods, respectively. In the next step, we construct and prune the retrieval tree. Then, we fuse the two trees, resulting in the retrieval-augmented draft token tree. Finally, the retrieval-augmented draft token tree is verified recursively. In the figure, green tokens denote $y_0$ in the current turn, and red tokens are accepted by the target model.}
   \label{fig:enter-label}
\end{figure*}

\subsection{Our Approach: RASD}
Our method RASD consists of three main steps: retrieval process, retrieval tree construction, and tree fusion. An overview of these steps is shown in Figure \ref{fig:enter-label}.

\begin{algorithm}
\caption{Retrieve Method of Suffix Match}
\label{1}
\begin{algorithmic}[1]
\REQUIRE 
\(x_n\) is the \(n\)-th token of the draft model input, and \(y_0\) is the first token generated by the target model. Datastore \(D\), result set \(S\), suffix length limits \(n_{max}\) and \(n_{min}\), number of candidates \(n\), and length of candidates \(l\). 

\FOR{$i = n_{max}$ \textbf{downto} $n_{min}$}
    \IF{\(\text{len}(S) \geq n\)}
        \STATE \textbf{break} 
    \ENDIF
    \STATE \(suffix \gets (x_{n-i+1}, \ldots, x_n, y_0)\) 
    
    \FOR{\textbf{each} \(i\_gram\) \textbf{in} \(D\)}
        \IF{\(i\_gram == suffix\)}
            \STATE \(c \gets\) next \(l\) tokens after \(i\_gram\) in \(D\)
            \IF{\(c   \textbf{ not in }   S\)}
                \STATE \(\text{Add } c \text{ to } S\) 
            \ENDIF
            \IF{\(\text{len}(S) \geq n\)}
                \STATE \textbf{break} 
            \ENDIF
        \ENDIF
    \ENDFOR
\ENDFOR
\end{algorithmic}
\end{algorithm}

\subsubsection{Retrieval Process}
In RASD, retrieval is employed to enhance the quality of candidate draft tokens by leveraging knowledge sources. We evaluate two retrieval settings: PLD and REST. To optimize retrieval efficiency and accuracy, an exact-match approach, inspired by PLD, is utilized to identify continuation candidates.

Given the context \(s^{\prime}\), we retrieve context-continuation pairs from the datastore \(D\), generating a set of continuation candidates \(S\):
\begin{align}
    S = \text{Retrieve}(D, s^{\prime}),
\end{align}
where \(\text{Retrieve}(D, s^{\prime})\) is a retrieve method based on suffix matching, which returns a set of retrieval results using \(s^{\prime}\) as the query to find contexts in \(D\) that match the longest suffix of \(s^{\prime}\). We illustrate a detailed process
in Algorithm \ref{1}. After this process, we obtain the candidate sequences.

For PLD, the retrieval process is designed to be iterative. If no results are initially found, we recursively select the top-$k$ tokens generated by the draft model and append them to the context \(s^{\prime}\). This process continues until retrieval results are successfully obtained. This iterative approach improves the success rate of retrieval, even when the initial query fails to yield matches. 

For REST, the retrieval process is limited to a single execution within each draft phase. This restriction is imposed due to the method's higher time overhead. This approach balances the richness of retrieved candidates with the practical constraints of runtime performance.

\subsubsection{Retrieval Tree Pruning}
We then construct the retrieval tree using the set of candidates \(S\). For PLD, each candidate \(t_i\) can be treated as a linked list. We merge nodes in \(S\) that share the same prefix, transforming these linked lists into one or more trees. The last token \(y_0\) of input \(s^{\prime}\) is used as the root node to merge these trees, resulting in the retrieval tree \(T_r\). For REST, the number of retrieval results can be substantial. To manage this, we construct the retrieval tree \(T_r\) by prioritizing candidates \(t_i\) using high-frequency prefixes as filters, following the same approach as REST.

Next, we propose a method to prune the retrieval tree using the draft model’s output distribution. There is a strong positive correlation between the draft model’s confidence score and the token acceptance rate. We leverage this confidence score to prune the retrieval tree.

In the first layer of target model output, the probability distribution is $P_1$:
\begin{align}
    P_1 = P(x \mid s^{\prime}; \theta_{\text{draft}}).
\end{align}

We hypothesize that the first token of a high-quality retrieval result should have a high probability of appearing in $P_1$. We reject the search results whose first token is not in the top-$k$ of $P_1$.

\subsubsection{Tree Fusion}
Thus far, we have obtained both the retrieval tree and the tree generated by the draft model. To produce the final output tree, we combine the draft model's generation tree with the retrieval tree.

Given that the forward propagation time of a large language model theoretically scales quadratically with input length, tree fusion can significantly reduce this time by eliminating redundant input tokens. Specifically, we perform a simple longest prefix match on each branch of both trees. Branches with identical prefixes in the two trees are merged, using the last node of the shared prefix as the parent node. This classic algorithm can be efficiently implemented using a trie tree.

After fusing the two trees, the attention matrix and position embeddings for the nodes of the new tree must be updated accordingly. The final combined tree integrates information from both the draft language model and knowledge base retrieval. Knowledge-based retrieval effectively addresses the out-of-domain problem and increases the upper limit of the length of candidate draft sequences.

\begin{table*}
  \centering
  \begin{tabular}{llccccccccccc}
    \hhline{=============}
    & & \multicolumn{2}{c}{\textbf{HumanEval}} & \multicolumn{2}{c}{\textbf{CNN/DM}} & \multicolumn{2}{c}{\textbf{MultiFieldQA}} & \multicolumn{2}{c}{\textbf{Qasper}} & \multicolumn{2}{c}{\textbf{DPR}} \\
    \hline
    \textbf{model} & \textbf{method} & SR & $\tau$ & SR & $\tau$ & SR & $\tau$ & SR & $\tau$ & SR & $\tau$   \\
    \hline
    \multicolumn{12}{c}{Temperature = 0} \\
    \hline
    \multirow{3}{*}{Q 14B} & PLD & 1.49 & 0.55 & 1.59 & 0.70 & 1.98 & 1.60 & 1.62 & 1.58 & 1.57 & 0.70 \\
     & EAGLE2 & 2.84 & 3.10 & 2.23 & 2.41 & 1.65 & 1.73 & 1.48 & 1.83 & 2.39 & 2.60  \\
     & RASD(PLD) & \textbf{2.98} & \textbf{3.44} & \textbf{2.31} & \textbf{2.68} & \textbf{2.10} & \textbf{2.78} & \textbf{1.66} & \textbf{2.58} & \textbf{2.43} & \textbf{2.92} \\
     \hline
    \multirow{3}{*}{L 8B} & PLD & 1.38 & 0.51 & 1.34 & 0.53 & 1.90 & 1.39 & 1.46 & 1.04 & 1.55 & 0.75\\
     & EAGLE2 & 2.76 & 3.37 & 2.15 & 2.51 & 1.68 & 2.28 & 1.52 & 2.36 & 2.37 & 2.97\\
     & RASD(PLD) & \textbf{2.79} & \textbf{3.55} & \textbf{2.21} & \textbf{2.72} & \textbf{1.98} & \textbf{2.86} & \textbf{1.59} & \textbf{2.84} & \textbf{2.44} & \textbf{3.38}\\
    \hline
    \multicolumn{12}{c}{Temperature = 1} \\
    \hline
    \multirow{3}{*}{Q 14B} & PLD & 1.53 & 0.58 & 1.42 & 0.49 & 1.72 & 1.25 & 1.54 & 1.53 & 1.74 &0.59\\
     & EAGLE2 & 2.73 & 3.09 & 1.95 & 2.03 & 1.56 & 1.61 & 1.39 & 1.76 & 2.40 & 2.36 \\
     & RASD(PLD) & \textbf{3.00} & \textbf{3.39} & \textbf{1.98} & \textbf{2.17} & \textbf{1.93} & \textbf{2.29} & \textbf{1.60} & \textbf{2.48} & \textbf{2.49} & \textbf{2.63} \\
     \hline
    \multirow{3}{*}{L 8B} & PLD & 1.32 & 0.45 & 1.37 & 0.43 & 1.81 & 1.45 & 1.53 & 0.95 & 1.42 &0.65\\
     & EAGLE2 & 2.57 & 3.38 & 2.04 & 2.26 & 1.60 & 2.35 & 1.53 & 2.61 & 2.19 & 2.76\\
     & RASD(PLD) & \textbf{2.59} & \textbf{3.48} & \textbf{2.13} & \textbf{2.42} & \textbf{1.82} & \textbf{2.88} & \textbf{1.66} & \textbf{3.16} & \textbf{2.24} & \textbf{3.11}\\
    \hhline{=============}
  \end{tabular}
  \caption{\label{citation-guide1}
    The speedup performance of our proposed RASD with PLD retrieval method and baselines in different datasets. We test on Qwen2.5-14B and LLaMA3-instruct-8B with temperatures of generation 0 and 1. The best results among all methods are in bolded.
  }
\end{table*}

\subsubsection{Draft Tree Verification}
We adopt a recursive verification strategy to accommodate tree attention. Utilizing tree attention, the target LLM calculates the probability of each token in the tree-structured draft within a single forward pass. Each node of the tree is arranged in one dimension, following the order of level-order traversal. The position embedding and attention matrices respectively represent the current node's level in the tree and its relationship with the parent node.

At each node of the draft tree, we recursively apply speculative decoding algorithms to sample or adjust the probability distribution, in line with the approach of SpecInfer \cite{DBLP:journals/corr/abs-2305-09781}.
\section{Experiments}

\subsection{Models and Tasks}
To evaluate the effectiveness of RASD in accelerating large language models, we use EAGLE2 as our draft model and conduct a series of experiments with two different target models across various tasks. We tested RASD on the LLaMA3-Instruct-8B \cite{DBLP:journals/corr/abs-2407-21783} and Qwen2.5-Instruct-14B \cite{DBLP:journals/corr/abs-2412-15115} models to assess its acceleration capabilities. The RASD (PLD) retrieval enhancement method is anticipated to perform well on tasks where the input contains potential output n-grams, such as in RAG \cite{DBLP:conf/emnlp/KarpukhinOMLWEC20}, DocQA, Summary, and Code tasks. Therefore, we selected the HumanEval, CNN/Daily Mail, MultiFieldQA, Qasper, and DPR datasets for our experiments. To push the boundaries of the RASD method, we utilize the REST-based retrieval method RASD (REST) to enhance tasks in more scenarios, choosing the HumanEval, NL2SQL, and Med-QA datasets for RASD (REST).

Both greedy sampling and non-greedy sampling are considered in all experiments to comprehensively evaluate speculative decoding performance. All evaluations are conducted on an NVIDIA A100 80G GPU.

\subsection{Metrics}
In our experiments, we adopt two primary metrics: average acceptance length $\tau$ and speedup ratio (SR). The average acceptance length $\tau$ evaluates the average number of tokens accepted per forward pass by the target large language models, excluding any overhead associated with retrieving or constructing draft tokens. This metric indicates the maximum possible acceleration. The second metric, speedup ratio (SR), measures the relative improvement in decoding speed compared to vanilla auto-regressive decoding.

\begin{table*}[h]
  \centering
  \begin{tabular}{llcccccc}
    \hhline{========}
    & & \multicolumn{2}{c}{\textbf{HumanEval}} & \multicolumn{2}{c}{\textbf{NL2SQL}} & \multicolumn{2}{c}{\textbf{MedQA}}  \\
    \hline
    \textbf{model} & \textbf{method} &  SR & $\tau$ & SR & $\tau$ & SR &$\tau$ \\
    \hline
    \multicolumn{8}{c}{Temperature = 0} \\
    \hline
    \multirow{3}{*}{Q 14B} & REST & 1.69 & 0.86 & 2.36 & 2.39 & 1.66 & 0.84  \\
     & EAGLE2 & 2.72 & 3.10 & 2.05 & 2.36 & 2.46 & 2.57   \\
     & RASD(REST) & \textbf{2.86} & \textbf{3.46} & \textbf{2.56} & \textbf{2.45} & \textbf{2.57} & \textbf{2.94}  \\
     \hline
    \multirow{3}{*}{L 8B} & REST & 1.70 & 0.84 & 3.59 & 1.51 & 1.85 & 0.89 \\
     & EAGLE2 & 2.76 & 3.36 & 3.78 & 2.76 & 3.13 & 2.76\\
     & RASD(REST) & \textbf{2.86} & \textbf{3.75} & \textbf{4.07} & \textbf{3.41} & \textbf{3.26} & \textbf{3.00} \\
    \hline
    \multicolumn{8}{c}{Temperature = 1} \\
    \hline
    \multirow{3}{*}{Q 14B} & REST & 1.76 & 0.83 & 2.32 & 2.19 & 1.42 & 0.50 \\
     & EAGLE2 & 2.73 & 3.09 & 2.07 & 2.35 & 2.24 & 2.06  \\
     & RASD(REST) & \textbf{2.90} & \textbf{3.48} & \textbf{2.40} & \textbf{3.30} & \textbf{2.35} & \textbf{2.36} \\
     \hline
    \multirow{3}{*}{L 8B} & REST & 1.53 & 0.85 & 3.58 & 1.14 & 1.30 & 0.51 \\
     & EAGLE2 & 2.57 & 3.38 & 4.05 & 2.77 & 2.24 & 2.47 \\
     & RASD(REST) & \textbf{2.59} & \textbf{3.44} & \textbf{4.24} & \textbf{3.26} & \textbf{2.35} & \textbf{2.75} \\
    \hhline{========}
  \end{tabular}
  \caption{\label{citation-guide2}
    The speedup performance of our proposed RASD with REST retrieval method and baselines in different datasets. We test on Qwen2.5-14B and LLaMA3-instruct-8B with temperatures of generation 0 and 1. The best results among all methods are in bolded.
  }
\end{table*}

\subsection{Baseline}
In this study, we focus solely on lossless speculative decoding approaches for LLMs. Among the methods that do not rely on draft models, we examine Prompt Lookup Decoding (PLD) \cite{saxena2023prompt}, REST \cite{DBLP:conf/naacl/0012ZCL024}, and EAGLE-2 \cite{DBLP:journals/corr/abs-2406-16858}, the latter being regarded as the state-of-the-art method for lossless speculative decoding tasks. Collectively, these baseline methods provide a robust framework for evaluating the efficiency of RASD in the LLM decoding process.

\subsection{Training}
We use the SharedGPT dataset, which comprises 68,000 dialogues from the Vicuna \cite{vicuna2023} series models' supervised fine-tuning dataset, as our training corpus. Given the substantial time and computational resources required, we choose not to regenerate responses for each dialogue turn using the target LLMs. Conducting training without re-generated data across all comparative methods remains equitable, although previous work \cite{DBLP:conf/icml/LiW0024} suggests that such an approach could slightly enhance the performance of the draft model. The learning rate is set to 5e-5, with ($\beta_1 = 0.9$, $\beta_2 = 0.95$
) for the AdamW \cite{DBLP:conf/iclr/LoshchilovH19} optimizer, and we implement gradient clipping at 0.5. We utilized eight NVIDIA A100 80G GPUs for the training process.

\subsection{Experimental Results}
Table \ref{citation-guide1} shows the performance of RASD (PLD) compared to other methods. RASD (PLD) consistently achieves the highest speedup across all tasks and models. Unlike the PLD method, RASD can generate draft tokens that reflect the language model distribution. Compared to EAGLE2, RASD retrieves more accurate draft tokens. When the PLD score is high, the improvement with RASD (PLD) is more significant. RASD (PLD) shows the most improvement in DocQA tasks (MultiFieldQA and Qasper) because these tasks usually contain repeated paragraphs in the input. In tasks with fewer repeated input segments, such as code and summary tasks, the improvement with RASD (PLD) is less pronounced.

Table \ref{citation-guide2} shows the performance of RASD (REST) against other methods. RASD (REST) outperforms in terms of speedup across all tasks and models. Compared with PLD, REST has a larger retrieval space and applies to more scenarios. Therefore, the improvement of RASD (REST) verifies RASD's compatibility with more retrieval methods. For tasks in the field of medical question answering, RASD can also achieve better results. We consider that for most knowledge-dependent tasks, RASD can effectively improve the speed of speculative decoding by building a contextual database in the domain.

\section{Ablation Study}

\begin{table}[h]
\centering
\begin{tabular}{lcc} 
\toprule
\textbf{Method}  &\textbf{SR} & \textbf{$\tau$} \\
\midrule
EAGLE2 &2.73 & 3.09 \\
RASD(REST) w/o \textbf{p}& 2.82 & 3.62 \\
RASD(REST) w/o \textbf{tf} &2.87 & 3.48 \\
RASD(REST)   &\textbf{2.90} & \textbf{3.48} \\
\bottomrule
\end{tabular}
\caption{The impact of pruning and tree fusion operations in RASD on the speedup performance in HumanEval.}
\label{table3}
\end{table}
\subsection{Pruning and Tree Fusion}
We investigated how retrieval tree pruning and tree fusion impact the results of RASD experiments. We conducted experiments on the qwen2.5-14b model using the HumanEval dataset with the REST method. As shown in Table \ref{table3}, both methods enhance the speedup ratio, with RASD incorporating pruning yielding a more pronounced effect. Pruning reduces the average acceptance length by eliminating unnecessary retrieval results early, thereby saving time required for the target model's forward pass. Tree fusion has a relatively smaller impact. It does not alter the average acceptance length but reduces redundant draft tokens, consequently saving time for the target model's forward pass.

\subsection{Length of Retrieval Candidates}
\begin{figure}[t!]
    \centering
    \begin{subfigure}[b]{0.23\textwidth}
        \centering
        \begin{tikzpicture}[scale=0.45]
            \begin{axis}[
                xlabel=Length of Retrieval Candidates,
                ylabel=Speedup Rate,
                grid=major,
                legend style={at={(0.5,-0.2)}, anchor=north, legend columns=-1} 
                ]
            \addplot[color=blue!50, line width=2pt, mark=square*, mark size=2.5pt] coordinates {
                (2,1.12)
                (4,1.37)
                (6,1.55)
                (8,1.62)
                (10,1.73)
                (12,1.98)
                (14,1.89)
                (16,1.90)
            };
            \addlegendentry{PLD}
            \addplot[color=red!60, line width=2pt, mark=triangle*, mark size=5pt] coordinates {
                (2,1.61)
                (4,1.87)
                (6,1.96)
                (8,2.00)
                (10,2.09)
                (12,2.10)
                (14,2.07)
                (16,2.02)
            };
            \addlegendentry{RASD(PLD)}
            \addplot[color=orange!80, line width=2pt, mark=star, mark size=5pt, dash dot] coordinates {
                (2,1.65)
                (16,1.65)
            };
            \addlegendentry{EAGLE2}
            \end{axis}
        \end{tikzpicture}
        \label{fig:rasdpld}
    \end{subfigure}
    \hfill
    \begin{subfigure}[b]{0.23\textwidth}
        \centering
        \begin{tikzpicture}[scale=0.45]
            \begin{axis}[
                xlabel=Length of Retrieval Candidates,
                ylabel=Speedup Rate,
                grid=major,
                legend style={at={(0.5,-0.2)}, anchor=north, legend columns=-1} 
                ]
            \addplot[color=blue!50, line width=2pt, mark=square*, mark size=2.5pt] coordinates {
                (2,1.28)
                (4,1.45)
                (6,1.58)
                (8,1.63)
                (10,1.66)
                (12,1.62)
                (14,1.59)
                (16,1.54)
            };
            \addlegendentry{REST}
            \addplot[color=red!60, line width=2pt, mark=triangle*, mark size=5pt] coordinates {
                (2,2.28)
                (4,2.39)
                (6,2.52)
                (8,2.57)
                (10,2.55)
                (12,2.52)
                (14,2.51)
                (16,2.48)
            };
            \addlegendentry{RASD(REST)}
            \addplot[color=orange!80, line width=2pt, mark=star, mark size=5pt, dash dot] coordinates {
                (2,2.46)
                (16,2.46)
            };
            \addlegendentry{EAGLE2}
            \end{axis}
        \end{tikzpicture}
        \label{fig:rasdrest}
    \end{subfigure}
    \caption{RASD performance with different lengths of retrieval candidates compared with the baselines}
    \label{fig:probknowledge}
\end{figure}
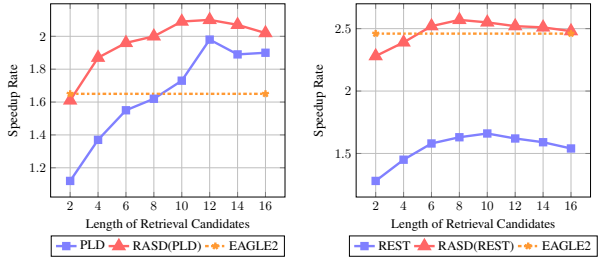

In the retrieval phase, the length of retrieval candidates $l$ is a crucial variable. If $l$ is small, the advantage of retrieval may not be significant, and if $l$ is large, the verification phase will require more time. We construct experiments on qwen2.5-14b, MultiFieldQA dataset used RASD(PLD) and qwen2.5-14b, MedQA dataset used RASD(REST) with different length of retrieval candidates $l$. As shown in \ref{fig:probknowledge}, On the left, since PLD is suitable for DocQA tasks, it can be seen that when only the PLD method is used, as $l$ increases, it can be inferred that the acceptance length of PLD is increasing. When \( l > 0 \), the performance exceeds EAGLE2. However, when $l$ continues to increase, the increase in verification time cost will limit the effect. RASD (PLD) also shows the same trend. When $l$ is only 2, the performance decreases compared to EAGLE2. This is because PLD requires additional time cost and does not bring enough length of retrieval candidates. On the right, since Medqa's database is relatively small, the search results using only REST are not as good as those of EAGLE2. However, RASD (REST) can still surpass EAGLE2 with the help of REST. RASD achieves the best results when $l$ is 8.

\subsection{Effect of Datastore size}
\begin{table}[H]
\centering
\begin{tabular}{lcccc} 
\toprule
\textbf{Method} & \textbf{Size} & \textbf{Time} & \textbf{SR} & \textbf{$\tau$} \\
\midrule
EAGLE2 & - & -     & 2.73 & 3.09 \\
RASD(REST)   & 0.9 GB & 0.2 ms & 2.76 & 3.22 \\
RASD(REST)  & 8.7 GB & 0.6 ms & 2.85 & 3.40 \\
RASD(REST)  & 27 GB  & 0.7 ms & 2.90 & 3.48 \\
\bottomrule
\end{tabular}
\caption{The speedup with different datastore sizes in RASD(REST).}
\label{table4}
\end{table}

We explored how datastore size affects RASD performance in REST. Increasing the datastore size improves the accuracy of retrieved draft tokens, which significantly boosts generation speed. As shown in Table \ref{table4}, both average acceptance length and speedup ratio improve with larger datastore sizes. However, the speedup growth is less pronounced than the increase in mean generated length. This difference may be due to the overhead of retrieving draft tokens. We assume that in industrial applications, there will be enough disk storage to build large data stores and enough CPU cores for fast retrieval. Therefore, there is still potential to achieve even faster speeds with a larger datastore.

\section{Conclusion}
In this work, we propose RASD: Retrieval-Augmented Speculative Decoding. We use retrieval methods to improve the quality of candidate sequences from draft models. Concretely, RASD improves the speedup of speculative sampling on the out-of-domain datasets that are difficult for the draft model to handle and improves the maximum output length of draft models. We develop a pruning method to select appropriate retrieval sequences and we fuse the sequence tree from the draft model with the retrieval tree, creating a final combined tree for verification. Experiments have proven that our method is effective and exhibits strong scalability.

\section{Limitations}
Based on our experiments and conclusions, we conclude some limitations of our work as follows:
\begin{itemize}
    \item In our experiments, we only performed the draft model generation and retrieval process serially. parallelism of the two processes is not considered, which will result in the acceleration ratio not reaching the theoretical maximum value.
    \item We only consider pruning the retrieval tree by controlling hyperparameters, which is not automatic and using the same hyperparameters for different conversation types is not the best choice.
\end{itemize}

\section*{Ethics Statement}
We hereby declare that all authors of this article are aware of and adhere to the provided ACL Code of Ethics and honor the code of conduct.
\paragraph*{Use of Human Annotations}

Human annotations are only used in methodological research at the beginning of the work, to assist in analyzing the feasibility of the proposed solution. Annotators consented to the use of data for research purposes. We ensure that the privacy of all annotators is protected throughout the annotation process, and all of them are adequately paid according to local standards. Human annotations are not applied during the evaluation of our method.
\paragraph*{Risks}

In this paper, all datasets are obtained from official sources. The datasets adopted have been anonymized and do not contain offensive information. However, we cannot guarantee that the datasets do not contain socially harmful or toxic language.

\bibliography{latex/acl_latex}

\end{document}